\documentclass{article}
\usepackage{arxiv}
\usepackage{amsmath}
\usepackage[utf8]{inputenc} 
\usepackage[T1]{fontenc}    
\usepackage{hyperref}       
\usepackage{url}            
\usepackage{booktabs}       
\usepackage{amsfonts}       
\usepackage{nicefrac}       
\usepackage{microtype}      
\usepackage{lipsum}
\usepackage{graphicx}
\usepackage{algorithm}
\usepackage{algpseudocode}
\usepackage{float}
\usepackage{textcase}
\usepackage{shorttoc}  
\usepackage{float}
\usepackage{amsmath}
\usepackage{algorithm}
\usepackage{algpseudocode}
\usepackage{pifont}
\usepackage{etoolbox}
\usepackage{enumerate}
\usepackage{amsthm}
\usepackage{amssymb,amsmath,amsthm}

\newtheorem{theorem}{Theorem}
\graphicspath{ {./images/} }

\title{Making Sigmoid-MSE Great Again: Output Reset Challenges Softmax Cross-Entropy in Neural Network Classification}

\author{
Kanishka Tyagi,\\
Department of Electrical Engineering\\
The University of Texas at Arlington \\
Arlington, TX, 76010 \\
\texttt{kanishka.tyagi@mavs.uta.edu} \\
  \And
Chinmay Rane \\
Department of Electrical Engineering\\
The University of Texas at Arlington \\
Arlington, TX, 76010 \\
\texttt{chinmay.rane@mavs.uta.edu}\\
  \And
Ketaki Vaidya \\
Oracle Corporation\\
Santa Clara, CA, 95054 \\
\texttt{ketaki.vaidya08@gmail.com}\\
  \And
Jeshwanth Challgundla \\
Department of Electrical Engineering\\
The University of Texas at Arlington \\
Arlington, TX, 76010 \\
\texttt{jeshwanth.challagundla@mavs.uta.edu}\\
  \And
Soumitro Swapan Auddy \\
Department of Electrical Engineering\\
The University of Texas at Arlington \\
Arlington, TX, 76010 \\
\texttt{auddy.soumitro@gmail.com}\\
  \And
Michael Manry \\
Department of Electrical Engineering\\
The University of Texas at Arlington \\
Arlington, TX, 76010 \\
\texttt{manry@uta.edu}\\
}

\begin{document}

\maketitle
\begin{abstract}
This study presents a comparative analysis of two objective functions, Mean Squared Error (MSE) and Softmax Cross-Entropy (SCE) for neural network classification tasks. While SCE combined with softmax activation is the conventional choice for transforming network outputs into class probabilities, we explore an alternative approach using MSE with sigmoid activation. We introduce the Output Reset algorithm, which reduces inconsistent errors and enhances classifier robustness. Through extensive experiments on benchmark datasets (MNIST, CIFAR-10, and Fashion-MNIST), we demonstrate that MSE with sigmoid activation achieves comparable accuracy and convergence rates to SCE, while exhibiting superior performance in scenarios with noisy data. Our findings indicate that MSE, despite its traditional association with regression tasks, serves as a viable alternative for classification problems, challenging conventional wisdom about neural network training strategies.

\end{abstract}


\section{Introduction}
Neural networks have emerged as a fundamental component of deep learning, enabling models to learn complex patterns from data through interconnected layers of nodes. Backed by the Universal Approximation Theorem \cite{hornik1989multilayer}, which establishes their ability to approximate any continuous function to arbitrary precision, these models have achieved significant success across diverse applications, including image recognition \cite{simonyan2014very}, language processing \cite{vaswani2017attention}, and strategy games \cite{vinyals2017starcraft}. Central to the performance of these models are objective functions, which guide the training process by quantifying the difference between predicted and actual outputs. Traditionally, the \textbf{Softmax Cross-Entropy (SCE)} loss has been the standard choice for classification tasks, effectively converting outputs into class probabilities.

Despite the widespread use of SCE, recent studies have explored alternative objective functions to address specific limitations such as sensitivity to class imbalance and noisy data. For instance, Focal Loss has been designed to prioritize hard-to-classify examples in object detection \cite{lin2017focal}, while Mean Squared Error (MSE) has remained the conventional choice for regression tasks \cite{krizhevsky2017imagenet}. However, there is a growing interest in investigating whether MSE, paired with a sigmoid activation function, can also be effective for classification tasks. Unlike the standard softmax approach, the sigmoid activation combined with MSE offers a different learning dynamic that could unify classification and regression within a single framework.

In addition to objective functions, optimization algorithms play a critical role in training neural networks. Techniques like the Adam optimizer \cite{kingma2014adam} and Stochastic Gradient Tree \cite{schmidt2017minimizing} have contributed to efficient convergence by adapting learning rates or introducing hierarchical parameter updates. Meanwhile, advancements in activation functions—such as the Rectified Linear Unit (ReLU) \cite{nair2010rectified}—have enabled models to overcome vanishing gradient issues and accelerate training. However, the interplay between objective functions and activation mechanisms has not been fully explored, particularly for non-standard combinations like MSE with sigmoid activation.

This paper aims to bridge this gap by examining the use of \textbf{MSE} as an alternative to \textbf{SCE} for classification tasks. We propose a unified approach that leverages the \textbf{Output Reset} algorithm to mitigate inconsistent errors, potentially enhancing the robustness of classifiers. The study focuses on the foundational linear classifier framework but anticipates extending these insights to deeper architectures. Through experiments on datasets such as \textbf{MNIST}, \textbf{CIFAR-10}, and \textbf{Fashion-MNIST}, we evaluate the comparative performance of MSE and SCE, highlighting scenarios where MSE with sigmoid activation offers advantages, particularly in handling noisy data. Our findings challenge established conventions in neural network training and open new avenues for integrating classification and regression paradigms within a single objective function. This work contributes to a broader understanding of how different objective functions impact learning dynamics and suggests practical implications for designing more robust and adaptable neural network models.

\section{Background}
    
In training neural networks, the choice of objective functions and optimization techniques plays a pivotal role in achieving accurate and efficient learning. For classification tasks, categorical cross-entropy has been the dominant loss function, minimizing the discrepancy between predicted and true class probabilities. While alternative loss functions like Focal Loss have emerged to address specific challenges such as class imbalance, Mean Squared Error (MSE) has primarily been associated with regression tasks, where it measures the average squared difference between predicted and actual values. Optimization techniques remain fundamental to effective neural network training. Methods such as gradient descent and its adaptive variants, particularly Adam \cite{kingma2014adam}, iteratively adjust model parameters to minimize the chosen objective function. These algorithms incorporate momentum, learning rate schedules, and adaptive adjustments to accelerate convergence and improve performance.

Nonlinear activation functions are crucial for capturing complex relationships within data. The Rectified Linear Unit (ReLU) \cite{nair2010rectified, jeshwanth2015thesis} and its variants have become standard choices, mitigating vanishing gradient issues and accelerating training. Recent research continues to explore adaptive combinations of activation functions to enhance network performance. Neural network optimization faces persistent challenges, including sensitivity to hyperparameter tuning, susceptibility to overfitting, and high computational demands. Novel approaches like Stochastic Gradient Tree (SGT) \cite{schmidt2017minimizing, state-based} introduce tree-based updates for parameter adjustments, offering improved convergence and robustness. Complementary methods in regularization and explainable AI seek to address overfitting and interpretability concerns. While categorical cross-entropy remains prevalent in classification tasks, exploring alternative objective functions offers promising avenues for innovation. This study investigates MSE paired with sigmoid activation for classification tasks, proposing a unified framework that bridges classification and regression paradigms \cite{multigain}. The sigmoid function traditionally provides probabilistic interpretation of outputs, but its combination with MSE introduces a novel learning dynamic that could impact classification performance.

To isolate the effects of this combination, we begin with a linear classifier analysis—providing a controlled environment to study the interaction between MSE loss and sigmoid activation. This foundational approach enables clear understanding of core principles before extending to more complex architectures, setting the groundwork for future research in deeper models.

\subsection{Structure and Notation}
\label{subsection:structure_and_notation}
    
\begin{figure}[h]
        \begin{center}
            \includegraphics[width=3.2in]{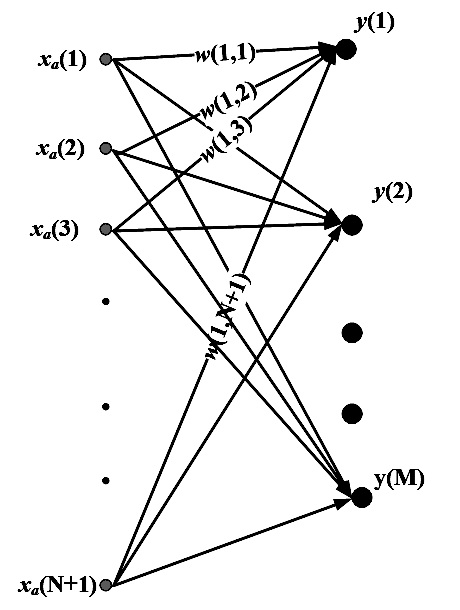}
        \end{center}
        \caption{Linear classifier}
        \label{fig:linear classifier}
    \end{figure}

Consider Fig. \ref{fig:linear classifier}, a linear classifier is represented by a weight matrix \(\mathbf{W}\), which transforms an input vector \(\mathbf{x}\) into a discriminant vector \(\mathbf{y}\) \cite{mostafa2012learning}. Each element \(w(m,n)\) in the weight matrix connects the \(n^{\text{th}}\) input to the \(m^{\text{th}}\) output. 
    
The training dataset is denoted as \((\mathbf{x}_p, \mathbf{t}_p)\), where \(\mathbf{x}_p\) is an \(N\)-dimensional input vector and \(\mathbf{t}_p\) is an \(M\)-dimensional target output vector. The index \(p\) ranges from 1 to \(N_v\), where \(N_v\) represents the total number of training patterns. To incorporate a threshold, the input vector \(\mathbf{x}\) is augmented with an additional element set to 1, resulting in the augmented input vector \(\mathbf{x}_a = [\mathbf{1} : \mathbf{x}^T]^T\). Thus, \(\mathbf{x}_a\) contains \(N_u\) basis functions, where \(N_u = N + 1\). For the \(p^{\text{th}}\) training pattern, the output vector \(\mathbf{y}_p\) is computed as:
    \begin{equation} \label{eq:linear_output}
        \mathbf{y}_p = \mathbf{W} \cdot \mathbf{x}_{ap}
    \end{equation}
    
    \noindent where \(\mathbf{x}_{ap}\) denotes the augmented input vector \(\mathbf{x}_a\) for the \(p^{\text{th}}\) pattern.  

    \subsection{Regression-Based Classifier}
    \label{subsection:regression_based_classifier}
    
Training the linear classifier involves minimizing an error function \(E\), which serves as a surrogate for a non-smooth classification error. Following the approach described in \cite{bishop2006pattern}, we adopt a Bayesian perspective, where training can be viewed as maximizing the likelihood function or minimizing the Mean Squared Error (MSE) in a least squares sense. The MSE between the predicted outputs and target outputs is defined as:

    \begin{equation}  
        E = \frac{1}{N_v}\sum_{p=1}^{N_v}\sum_{i=1}^{M}[t_p(i) - y_p(i)]^2 
        \label{eq:error}
    \end{equation}
    
Here, the target output for the correct class \(i_c\) is denoted by \(t_p(i_c) = b\), while for all incorrect classes \(i_d\), \(t_p(i_d) = 0\), where \(b\) is a positive constant, typically set to 1. \(M\) represents the total number of classes, and a one-versus-all coding scheme is employed. Minimizing the error function in equation (\ref{eq:error}) with respect to \(\mathbf{W}\) requires solving \(M\) sets of \(N + 1\) linear equations, expressed as:

    \begin{equation} 
        \mathbf{C} = \mathbf{R \cdot W^T}
        \label{eq:ols-eq}
    \end{equation}
    
In this context, \(\mathbf{C}\) is the cross-correlation matrix and \(\mathbf{R}\) is the auto-correlation matrix, defined respectively as:

    \begin{equation}
        \mathbf{C} = \frac{1}{N_v}\sum_{p=1}^{N_v} \mathbf{x}_{ap} \cdot \mathbf{t}_p^T  
        \label{eq:cross_auto_correlation}
    \end{equation}
    
    \begin{equation}
        \mathbf{R} = \frac{1}{N_v}\sum_{p=1}^{N_v}\mathbf{x}_{ap} \cdot \mathbf{x}_{ap}^T  
        \label{eq:auto_correlation}
    \end{equation}
    
However, due to potential ill-conditioning of \(\mathbf{R}\), solving equation (\ref{eq:ols-eq}) using Gauss-Jordan elimination is not recommended. Instead, Orthogonal Least Squares (OLS) \cite{ols} is employed, as it offers a more stable solution. OLS has two main advantages: it enables fast training by solving linear equations efficiently, and it helps avoid some local minima \cite{ols2}. In optimization terms, solving equation (\ref{eq:ols-eq}) for \(\mathbf{W}\) corresponds to applying Newton's algorithm for updating the output weights \cite{melvinrobinson2013}.

The linear output activation functions as a class discriminant. The classifier is considered to correctly classify the \(p^{\text{th}}\) pattern when the output \(y_p(i_c)\) is the highest among all classes, where \(i_c\) is the correct class index. Let \(i_c^{'}\) denote the estimated class for a given pattern, defined as:

    \begin{equation}
        i_{c}^{'} = \underset{i}{\text{arg max}} \; y_p(i)
    \end{equation}
    
If \(i_c^{'} = i_c\), the network correctly classifies the pattern; otherwise, it is misclassified. It should be noted that errors in regression-based classifiers are unbounded, as they can satisfy \(-\infty < y_p(i) - t_p(i) < \infty\).

\section{Mathematical Background}
    
In neural networks, given an input vector \(\mathbf{x}\), the network computes a set of scores for each class \(c\), represented as \(z_c\). These scores estimate the posterior probabili ties of classes, denoted by \(q(c|\mathbf{x})\), which are typically converted to probabilities using the softmax function:
    
\[
 q(c|\mathbf{x}) = \frac{e^{z_c}}{\sum_{c'} e^{z_{c'}}}
 \]
    
For simplicity, we use \(q(c|\mathbf{x})\) without explicitly including the model parameters \(\theta\). The cross-entropy loss measures the difference between the predicted class probabilities and the true class labels, defined as:
    
\[
\text{CE} = -\sum_{n=1}^{N} \sum_{c} \delta(c, c_n) \log(q(c|\mathbf{x}_n))
\]
    
Here, \(N\) represents the number of training examples, \(c\) iterates over all possible classes, \(\delta(c, c_n)\) is the Kronecker delta function (1 if \(c = c_n\), 0 otherwise), and \(\mathbf{x}_n\) and \(c_n\) denote the \(n\)-th training example and its true class label, respectively. Cross-entropy loss drives the network to produce accurate class probability estimates. From a Maximum Likelihood Estimation (MLE) perspective, minimizing the cross-entropy loss is equivalent to maximizing the likelihood of the observed data under the model. The likelihood \(P(D|\theta)\), where \(D\) represents the dataset and \(\theta\) the model parameters, can be expressed as:
    
    \[
    P(D|\theta) = \prod_{n=1}^{N} q(c_n|\mathbf{x}_n)
    \]
    
Thus, maximizing the likelihood is equivalent to minimizing the cross-entropy loss, making it an MLE-based approach to training neural networks. The Mean Squared Error (MSE) measures the squared difference between predicted probabilities and true labels:
    
    \[
    \text{MSE} = \frac{1}{N} \sum_{n=1}^{N} \sum_{c} (q(c|\mathbf{x}_n) - \delta(c, c_n))^2
    \]
    
Similar to cross-entropy, MSE can be understood from a Bayesian perspective. Assuming a Gaussian likelihood model, minimizing MSE corresponds to maximizing the posterior probability of the model parameters \(\theta\) given the data \(D\), aligning with Maximum A Posteriori (MAP) estimation:
    
    \[
    P(\theta|D) \propto P(D|\theta) \cdot P(\theta)
    \]
    
Here, \(P(\theta)\) represents the prior distribution over the model parameters. In this context, minimizing MSE is analogous to maximizing the posterior probability under the Bayesian framework. To analyze convergence, we examine how gradient computations affect convergence under different loss criteria: cross-entropy (CE) and mean squared error (MSE). Both criteria relate to estimating class posterior probabilities during neural network training. Cross-entropy aligns with MLE, while MSE is associated with MAP estimation, especially when a Gaussian likelihood is assumed. A neural network's parameters are denoted as \(\theta = \{w_l \in \mathbb{R}\}\). The output of each neuron is computed by applying nonlinear activation functions to the linear combination of connected inputs. Ignoring bias terms, the input to the final layer is calculated as \(z_c = \mathbf{w} \cdot \mathbf{y}^{L-1}\). For linear classifiers, \(\mathbf{y}^{L-1} = \mathbf{x}\). The final output \(y_c^L\) is obtained by applying the softmax transformation: \(y_c^L = \sigma(z_c) = q_\theta(c|\mathbf{x}_n)\). Given a training set of labeled samples \(\{(\mathbf{x}_n, c_n) : 1 \leq n \leq N\}\), the goal is to minimize the global error function:
    
    \[
    E_{\text{global}} = \frac{1}{N} \sum_{n} E(q_{\theta}(\cdot|\mathbf{x}_n, c_n))
    \]
    
Typically, this minimization is achieved using stochastic gradient descent (SGD), with weight updates performed in the direction of the negative gradient. The gradient of the error with respect to weight parameters, \(\frac{\partial E}{\partial w_l}\), is given by:
    
    \begin{equation}
    \frac{\partial E}{\partial w_l} = \Delta_l \cdot y^{(l-1)}
    \end{equation}
    
The error signal \(\Delta_L\) in the final layer depends on both the loss criterion and the activation function. For cross-entropy with softmax, the error signal is:
    
    \begin{equation}
    \Delta_{CE}(y_c) = y_c - \delta(c, c_n)
    \end{equation}
    
Analyzing the functional dependencies, the error signal for the cross-entropy criterion can be expressed as:
    
    \begin{equation}
    \Delta_{CE}(y_c) =
    \begin{cases}
    y_c - 1 & \text{if } c = c_n \\
    y_c & \text{if } c \neq c_n
    \end{cases}
    \end{equation}
    
For \textbf{MSE with sigmoid activation}, the error signal is defined as:
    
    \begin{equation}
    \Delta_{MSE}(y_c) = -2 \sum_{i=1}^{M} [t_i - y_i] \cdot y_c (1-y_c)
    \end{equation}
    
    where \(t_i\) is defined as:
    
    \begin{equation}
    t_i =
    \begin{cases}
    1 & \text{if } c = c_n \\
    0 & \text{if } c \neq c_n
    \end{cases}
    \end{equation}
    
\subsection{Problems With MSE Type Training of Neural Networks}

Current regression-based classifiers, as well as some other models, face several challenges that hinder their performance. Our goal is to minimize the probability of testing error $Pe_{tst}$ in a multi-class application where $M$ = $N_c$ $\geq$ $2$. Before we do this, let’s consider problems that make it difficult to
minimize the training MSE. 

\begin{enumerate}[({P}1)]
 
 \item \textbf{Pattern bias errors:} where the mean $m_t(p)$ of $t_p(i)$ differs from the mean $m_y(p)$ of $y_p(i)$ increase $E$ but not $P_{et}$.

 \item \textbf{Consistent errors:} The errors \(y_p(i) - t_p(i)\) may also be consistent, with their absolute values moving in the same direction as \(P_e\). This arises when \(y_p(i_c) \le t_p(i_c)\) or \(y_p(i_d) \ge t_p(i_d)\) for any incorrect class \(i_d\).
        
 \item \textbf{Inconsistent errors:} The errors \(y_p(i) - t_p(i)\) can be inconsistent, with their absolute values moving in a direction opposite to that of the probability of classification error \(P_e\), while increasing the MSE. This typically occurs when \(y_p(i_c) \ge t_p(i_c)\) for the correct class \(i_c\) or \(y_p(i_d) \le t_p(i_d)\) for any incorrect class \(i_d\).

\item \textbf{Outliers:} Small consistent or inconsistent errors can escalate into outliers, which severely degrade performance \cite{tyagi2019multi}.
        
\item \textbf{Redundant inputs:} Excessive or irrelevant inputs lead to increased testing errors, a phenomenon often referred to as the Hughes phenomenon \cite{tyagi2019multi}, commonly known as overfitting in machine learning. Redundant inputs can be categorized as:
        \begin{enumerate}[({C}1)]
            \item Inputs that are linearly dependent on other inputs.
            \item Inputs that are independent but provide no useful information for calculating the output discriminant vector \(\mathbf{y}\).
        \end{enumerate}

\item \textbf{Unbound Error} Unlike for the Pe objective function, both
consistent errors and inconsistent errors are unbounded, so E is unbounded as seen
in the following lemma.

\end{enumerate}

$\textbf{\textit{Lemma 1}}$: Let $p$ denote an arbitrary training pattern for a neural net classifier. Then 
$\lim_{|y_p(i)| \to \infty} E \to \infty$

The objective function \( P_e \) has certain properties as follows:

\begin{enumerate}
    \item $P_e$ is bounded.
    \item  Each classification error counts as $\frac{1}{N_v}$, independent of:
    \begin{enumerate}
        \item  The amount by which $y(i_d)$ exceeds $y(i_c)$,
        \item  The difference between $m_y(p)$ and $m_t(p)$.
    \end{enumerate}
\end{enumerate}

\begin{figure}[h]
        \begin{center}
            \includegraphics[width=3.2in]{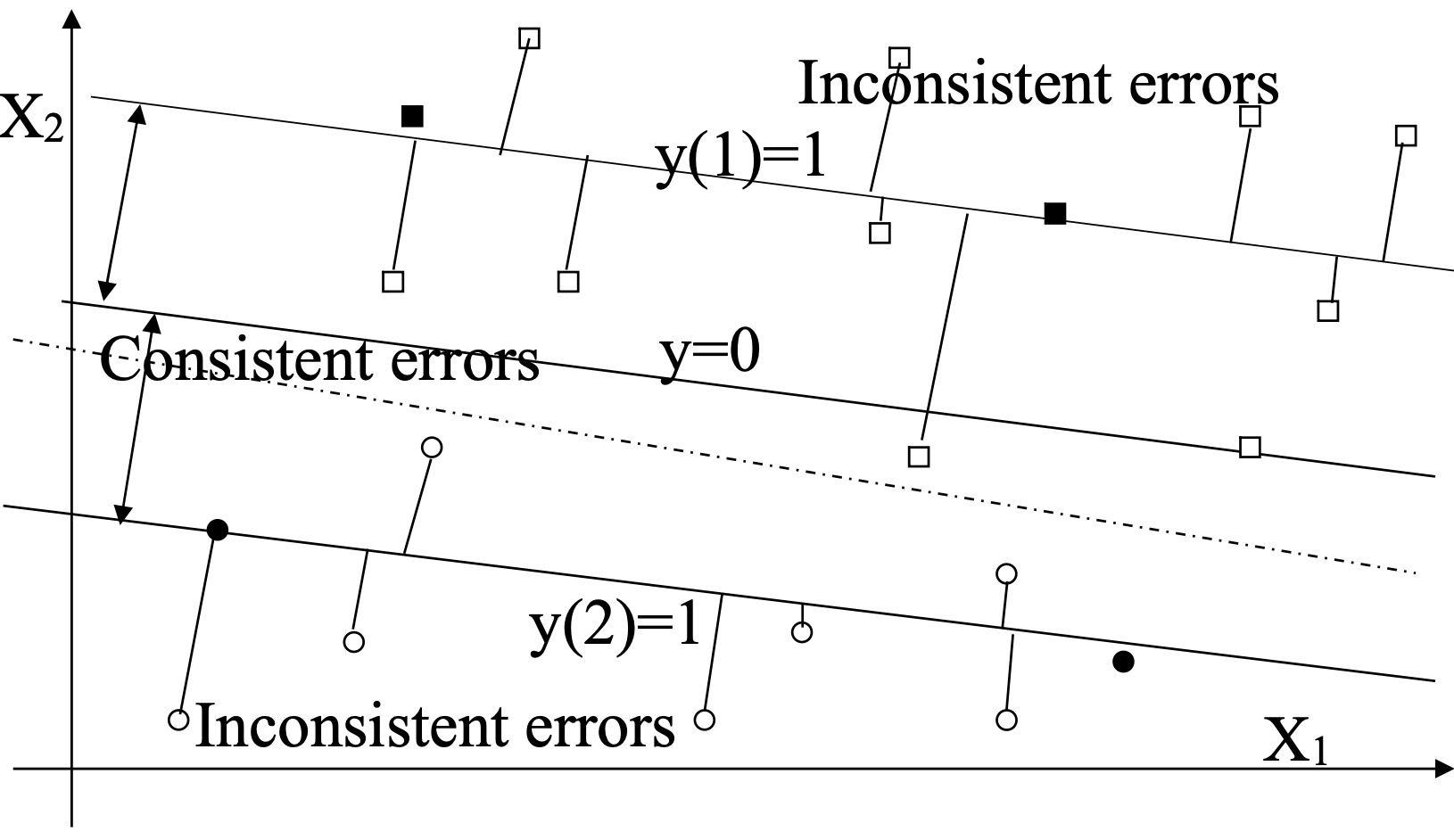}
        \end{center}
        \caption{MSE training example with \( N = M = N_c = 2 \), showing class boundary, optimal hyperplane, and support vectors. Squares and circles represent class patterns, with errors indicated by lines perpendicular to the boundary.}

        \label{fig:or1}
    \end{figure}

As an example in Figure \ref{fig:or1}, let \( N = M = N_c = 2 \), where \( t_p(i_c) = 1 \) and \( t_p(i_d) = 0 \). The upper and lower parallel lines represent \( y_p(i_c) = 1 \) and \( y_p(i_d) = 0 \), respectively. The middle line denotes the class boundary, and the dotted line represents an "optimal" separating hyperplane. Squares indicate patterns from class 1, while circles represent patterns from other classes. Dark circles and squares indicate support vectors where $y_p$ = $t_p$ on the margin hyperplanes. Errors are the straight lines perpendicular to the class boundary.

While margin-based classifiers are generally less affected by issues (P1) through (P6), as they primarily rely on support vectors, they often incorporate slack variables to minimize the impact of imperfect outputs. However, even these classifiers are still vulnerable to problem (P5) due to their inability to effectively eliminate redundant inputs. Notably, problem (P1) to (P4) are particularly relevant for Bayes-Gaussian and regression-based classifiers, which typically assume no presence of outliers. The proposed algorithm in this paper, however, operates without any assumptions about data statistics and is robust to outliers. Similarly, \cite{bishop2006pattern} introduces the \textbf{V matrix method}, which captures the geometric characteristics of the observation data, often overlooked by classical statistical methods, resulting in a smoother approximation.

\section{The Potential for MSE-Based Optimal Classifiers}

The minimum MSE solution for weight matrix $\mathbf{W}$ is obtained by solving equation \ref{eq:ols-eq}. Let $\mathbf{W}_{\text{opt}}$ denote the weight matrix for a classifier (a linear classifier in our case) that is optimal in some sense, such as the minimum probability of error (MPE) or Bayes classifier, or a classifier solving problems P1 through P6.
\begin{theorem}
$\mathbf{W}_{\text{opt}}$ is the solution to a least squares problem.
\end{theorem}

Given $\mathbf{W}{\text{opt}}$, the optimal cross-correlation matrix $\mathbf{C}{\text{opt}}$ is found from equation (1) as:
\begin{equation}
\mathbf{C}{\text{opt}} = \mathbf{R} \cdot (\mathbf{W}{\text{opt}})^T
\label{eq:copt}
\end{equation}
In equation \ref{eq:copt}, the basis vectors $\mathbf{X}p$ are given and unchangeable, so $\mathbf{R}$ is fixed. The only component of $\mathbf{C}{\text{opt}}$ under user control is the desired output vector $\mathbf{t}$. Therefore, the optimal cross-correlation matrix is found as:
\begin{equation}
\mathbf{C}{\text{opt}} = E[\mathbf{X} \cdot (\mathbf{t}{\text{opt}})^T]
\label{eq:copt2}
\end{equation}
Various classifiers with desirable properties can be designed through regression by properly choosing the desired output $\mathbf{t}$. Equation \ref{eq:copt2} can be rewritten as:
\begin{equation}
\mathbf{C}{\text{opt}} = \frac{1}{N_v}\sum{p=1}^{N_v} \mathbf{X}_p (\mathbf{t}_p)^T
\end{equation}

And further as:
\begin{equation}
\mathbf{c}(i) = \mathbf{A} \cdot \mathbf{t}(i)
\label{eq:ci}
\end{equation}
where $\mathbf{c}(i)$ is the $i^{\text{th}}$ column of $\mathbf{C}_{\text{opt}}$. The $(n,p)$ element of the $N_u \times N_v$ matrix $\mathbf{A}$ is $a(n,p) = X_p(n)/N_v$. The $p^{\text{th}}$ element of the column vector $\mathbf{t}(i)$ is $t_p(i)$. Assuming that $N_u \leq N_v$ and that $\mathbf{A}$ has rank $N_u$, equation \ref{eq:ci} represents an underdetermined set of equations for $\mathbf{t}(i)$ and has uncountably many exact solutions. The challenge lies in finding one without prior knowledge of $\mathbf{W}{\text{opt}}$ or $\mathbf{C}{\text{opt}}$. \textbf{Therefore, we need better ways for generating $t_p(i)$.}

\subsection{Solution for Problem P1 and $N_c = 2$}
Output reset (OR) \cite{tyagi2019multi} is used as an algorthm so solve the problem in P1. Use coded outputs so $t_p = t_p(1) - t_p(2)$ and  

\[
y_p = y_p(1) - y_p(2) = [t_p(1) + a_p] - [t_p(2) + a_p] = t_p(1) - t_p(2) 
\]
Ignoring other types of noise. The $N_c = 2$ coded output case is immune to problem P1, unlike the uncoded case. Also, both coded and uncoded cases for $N_c = 2$ can have the same $P_e$. What happens for $N_c > 2$?

\subsection{Solution for Problem P3}  
Extending the idea from \cite{tyagi2019multi}, this is similar to Ho-Kashyap \cite{fukunaga2013introduction} training. OR type 2 calculates $t'_{p}(i)$ as follows:
\begin{enumerate}
    \item If $y_p(i_c) > t_p(i_c)$ we set $t'_{p}(i_c) = y_p(i_c)$.
    \item If $y_p(i_d) < t_p(i_d)$ for an incorrect class $i_d$, we set $t'_{p}(i_d) = y_p(i_d)$.
\end{enumerate}

In each OWO-BP iteration,
\begin{enumerate}
    \item \textbf{BP Step:} Calculate $t'_{p}(i)$ and use it in the delta function calculations.
    \item \textbf{OWO Step:} Re-calculate $t'_{p}(i)$ and use $X_p \cdot (t'_{p})^T$ to update the $C$ matrix.
\end{enumerate}

Refer to \cite{tyagi2019multi} for details on OWO-BP. OR type 2 clearly eliminates the inconsistent errors of (P3) and can greatly decrease training and validation errors [17-20]. Note that for a classifier having zero classification errors, $E'$ will approach zero but $E$ will not. Therefore, training halts when the classifier has zero empirical risk $E'$. Typically, we use $t_p(i_c ) = +1$ and $t_p(i_d ) = 0$.

$\textbf{\textit{Lemma 2}}$: Let $p$ denote an arbitrary training pattern for a neural net classifier, trained using OR 2. Let $E_o$ denote the normal value of $E'$ when $y_p(i_c) = t_p(i_c)$ and $y_p(i_d) = t_p(i_d)$. Then  

\[
\lim_{y_p(i_c) \to +\infty} E' \to E_o \quad \text{and} \quad \lim_{y_p(i_d) \to -\infty} E' \to E_o
\]

\[
\lim_{y_p(i_c) \to -\infty} E' \to \infty \quad \text{and} \quad \lim_{y_p(i_d) \to +\infty} E' \to \infty
\]

\begin{enumerate}
    \item  $E'$ is well-behaved for patterns with inconsistent errors, but 
    \item  It still has problems P2 and P4.
    \item  OR type 2 does not minimize $P_e$.
\end{enumerate}

\subsection{Solution for P1, P2 and P4}  
An improved OR is to change $t_p(i)$ so that $E' \sim P_e$  

\textbf{Case 1:} If $x_p$ is correctly classified, $t'_p(i) = y_p(i)$ for all $i$, so no error results.

\textbf{Case 2:} If $x_p$ is misclassified and $y_p(i_d) > y_p(i_c)$ for one value of $i_d$, then $t'_p(i_c) = y_p(i_c) + b$, $t'_p(i_d) = y_p(i_d) - b$. Else, $t'_p(i) = y_p(i)$

Note that this pattern contributes $2b^2$ to $E'$, before the division by $N_v$.

\textbf{Case 3:} If $x_p$ is misclassified and $y_p(i_d) > y_p(i_c)$ for $K$ values of $i_d$, then $t'_p(i_c) = y_p(i_c) + K \cdot \epsilon$ and $t'_p(i_d(k)) = y_p(i_d(k)) - \epsilon$ for $1 \leq k \leq K$ where  

\[
\epsilon = b \sqrt{\frac{2}{K^2 + K}}
\]
Else, $t'_p(i) = y_p(i)$




\section{Enhancements in Linear Classifier Design}

This section builds upon the linear classifier \(\mathbf{W}\) described in subsection \ref{subsection:regression_based_classifier} and introduces a series of incremental improvements aimed at addressing the issues identified in (P1) through (P6). These enhancements involve modifications to the elements of the matrices \(\mathbf{R}\) and \(\mathbf{C}\) in equation (\ref{eq:ols-eq}).
    
\subsection{Target Adjustment using Output Reset}

The objective of this subsection is to eliminate the inconsistent errors identified in (P1) and reduce variations in output vectors \(\mathbf{y}_p\) caused by (P5). This is achieved by developing new target outputs, denoted as \(t_p^{'}(i)\), while maintaining the constraint that the target margin satisfies \(t_p^{'}(i_c) - t_p^{'}(i_d) \geq 1\). Inconsistent errors can increase the error function \(E\) while either decreasing the classification error \(P_e\) or leaving it unchanged. Two sources of inconsistent errors are considered: 

1. Bias in the outputs, where the average of \(t_p(i)\) differs from the average of \(y_p(i)\).
2. Situations where \(y_p(i_c) > t_p(i_c)\) or \(y_p(i_d) < t_p(i_d)\), as described in (P1). To mitigate these inconsistent errors, we introduce a new error function \(E^{'}\) \cite{or} \cite{or-gore}, where only the targets are adjusted, not the labels \cite{golub2012matrix}. Following the approach outlined in \cite{fukunaga2013introduction}, the new error function is defined as:
    
    \begin{equation}
        {E^{'} = \frac{1}{N_v}\sum_{p=1}^{Nv}\sum_{i=1}^{M}[t^{'}_p(i) - y_p(i)]^2}
        \label{eq:error_or}
    \end{equation}
    
    Here, the adjusted target \(t_p^{'}(i)\) is expressed as:
    \begin{equation}
        t^{'}_p(i)  = t_p(i) + a_p + d_p(i)
        \label{eq:tprime}
    \end{equation}
    
    where \(a_p\) and \(d_p(i)\) are initially set to zero. Since \(a_p\) is uniform across all classes, it has no effect on the classification error \(P_e\). By setting \(\frac{\partial E^{'}}{\partial a_p} = 0\), we derive the closed-form solution for \(a_p\), given by:
    
    \begin{equation}
        a_p = \frac{1}{M}\sum_{i=1}^{M}[y_p(i) - t_p^{'}(i) - d_p(i)]
        \label{eq:a_p}
    \end{equation}
    
    Similarly, \(d_p(i)\) is defined as:
    \begin{equation}
        d_p(i) = y_p(i) - t_p^{'}(i) - a_p 
    \end{equation}
    with the constraints that \(d_p(i_c) \geq 0\) and \(d_p(i_d) \leq 0\). While \(a_p\) and \(d_p(i)\) can be incorporated into \(t_p^{'}(i)\) or \(y_p(i)\) during training, they cannot be used during testing, as the correct class \(i_c(p)\) is unknown. Therefore, these adjustments are applied solely to \(t_p^{'}(i)\).
    
    To ensure the elimination of inconsistent errors, we require that \(t_p^{'}(i_c) \geq y_p(i_c)\) and \(t_p^{'}(i_d) \leq y_p(i_d)\), leading to:
    
    \begin{equation}
        d_p(i_c)  = [y_p(i_c) - a_p - t_p(i_c)] u(y_p(i_c) - a_p - t_p(i_c))
        \label{eq:dpic}
    \end{equation}
    
    \begin{equation}
        d_p(i_d)  = [y_p(i_d) - a_p - t_p(i_d)] u(t_p(i_d) - a_p - y_p(i_d))
        \label{eq:dpid}
    \end{equation}
    
    where \(u(\cdot)\) denotes the unit step function. The training halts when \(E^{'}\) becomes zero, even if the original error \(E\) is not zero. This approach reduces the impact of problem (P5), as \(\lim_{y_p(i_c) \to -\infty} (y_p^{'}(i_c) - t_p(i_c)) = 0\) and \(\lim_{y_p(i_d) \to \infty} (y_p^{'}(i_d) - t_p(i_d)) = 0\). We refer to this process as the \textbf{Output Reset (OR) algorithm}, described as follows:
    
    \begin{algorithm}[H]
        \caption{Output Reset (OR) Algorithm}
        \label{algo:or}
        \begin{algorithmic}[1]
            \State \textbf{Input:} \(p\), \(t_p(i)\), \(y_p(i)\)
            \State Initialize \(a_p = 0\), \(d_p(i) = 0\) for \(1 \leq i \leq M\)
            \For{\(i_t = 1\) to 3}
            \State Calculate \(a_p\) using equation (\ref{eq:a_p})
            \For{\(i = 1\) to \(M\)}
            \State Calculate \(d_p(i)\) using equations (\ref{eq:dpic}) and (\ref{eq:dpid})
            \State Update \(t_p^{'}(i)\) using equation (\ref{eq:tprime})
            \EndFor
            \EndFor
        \end{algorithmic}
    \end{algorithm}
    
    In this algorithm, limiting the maximum number of iterations to 3 allows significant performance improvement without a substantial increase in training time \cite{liu1994image}. In \cite{or-gore}, the heuristic iterative OR algorithm from \cite{liu1994image} was replaced with a closed-form solution for target outputs. Using the OR method, the training of regression-based linear classifiers (LC) is modified as follows:
    
    \begin{algorithm}[H]
        \caption{LC-OR Algorithm}
        \label{algo:lc_or}
        \begin{algorithmic}[1]
            \State Read the training data and set \(N_{it}^{1} = 10\)
            \State Calculate \(\mathbf{R}\) and \(\mathbf{C}\) using equation (\ref{eq:cross_auto_correlation})
            \State Solve equation (\ref{eq:ols-eq}) for the initial weight matrix \(\mathbf{W}\)
            \State Initialize \(i_t = 0\)
            \While{\(i_t < N_{it}^{1}\)}
            \State Set \(\mathbf{C} \leftarrow 0\)
            \For{\(p = 1\) to \(N_v\)}
            \State Apply the OR algorithm to transform \(\mathbf{t}_p\) into \(\mathbf{t}_p^{'}\)
            \State Update \(\mathbf{C}\) using \(\mathbf{C} \leftarrow \mathbf{C} + \mathbf{x}_{ap} (\mathbf{t}_p^{'})^T\)
            \EndFor
            \State Solve equation (\ref{eq:ols-eq}) for \(\mathbf{W}\)
            \State Increment \(i_t\) by 1
            \EndWhile
        \end{algorithmic}
    \end{algorithm}
    
    The OR method successfully addresses inconsistent errors (P1) and reduces the impact of output biases and outliers (P5) by adjusting the cross-correlation matrix \(\mathbf{C}\).
    
    \par An alternative approach to target modification is the \textbf{Ho-Kashyap algorithm} \cite{ho1965algorithm} \cite{ho1966class}. Unlike the OR method, Ho-Kashyap (1) does not account for pattern biases, (2) is applied with binary or one-versus-one coding, and (3) solves linear equations for both target vectors and classifier weights simultaneously.

    \section{Experimental Results}
    \label{sec:Experimental Results}
    
In this section, we present a comparative analysis of the 10-fold testing performance of three different linear classifiers: the \textbf{softmax cross-entropy (SCE)} classifier, the \textbf{mean squared error with output reset (MSE-OR)} classifier, and the \textbf{sigmoidal mean squared error with output reset (SMSE-OR)} classifier. The comparison focuses on the performance of the three classifiers, where each classifier is trained using the optimal learning factor optimization algorithm. The implementation details for these classifiers are described in the earlier sections. All models are trained for a single iteration, and 10-fold testing is employed to evaluate their performance. In this approach, the dataset is randomly partitioned into ten subsets of nearly equal size. In each fold, one subset is used for testing, seven subsets for training, and the remaining two subsets for validation. Before training, image data are normalized to have values between 0 and 1.
    
The models are trained for 5000 iterations, with accuracy evaluated based on the weights that yield the highest validation accuracy. The results for each classifier across various datasets are shown in Tables \ref{table:SCE_kfold_results}, \ref{table:MSE-OR_kfold_results}, and \ref{table:SMSE-OR_kfold_results}.

    \begin{table}
        \centering
        \caption{10-fold Testing Results for SCE Algorithm}
        \label{table:SCE_kfold_results}
        \begin{tabular}{|l|l|l|l|}
            \hline
            Dataset  & Average Testing PE & Best Average Validation Iteration \\ \hline
            MNIST & 8.0233 & 4861.8\\
            \hline
            CIFAR10 & 58.6850 & 3982.4\\
            \hline
            SVHN & 73.6124 & 4775.1\\
            \hline
            FASHION MNIST & 14.6671 & 4865.6\\
            \hline
            SCRAP DATA & 31.2678 & 3646.8\\
            \hline
        \end{tabular}
    \end{table}

    \begin{table}
        \centering
        \caption{10-fold Testing Results for MSE-OR Algorithm}
        \label{table:MSE-OR_kfold_results}
        \begin{tabular}{|l|l|l|l|}
            \hline
            Dataset  & Average Testing PE & Best Average Validation Iteration \\ \hline
            MNIST & 8.0517 & 1428.5\\
            \hline
            CIFAR10 & 59.6683 & 2271.7\\
            \hline
            SVHN & 74.4050 & 3903.7\\
            \hline
            FASHION MNIST & 14.5357 & 3916\\
            \hline
            SCRAP DATA & \textbf{31.1343} & 3757.9\\
            \hline
        \end{tabular}
    \end{table}
    
    \begin{table}
        \centering
        \caption{10-fold Testing Results for SMSE-OR Algorithm}
        \label{table:SMSE-OR_kfold_results}
        \begin{tabular}{|l|l|l|l|}
            \hline
            Dataset  & Average Testing PE & Best Average Validation Iteration \\ \hline
            MNIST & \textbf{7.9383} & 421.5\\
            \hline
            CIFAR10 & \textbf{58.1883} & 1493.3\\
            \hline
            SVHN & \textbf{61.1568} & 3769.9\\
            \hline
            FASHION MNIST & \textbf{14.2086} & 1384.1\\
            \hline
            SCRAP DATA & 38.1431 & 4934.3\\
            \hline
        \end{tabular}
    \end{table}
    
 Tables \ref{table:SCE_kfold_results}, \ref{table:MSE-OR_kfold_results}, and \ref{table:SMSE-OR_kfold_results} present the 10-fold testing results for each classifier. The results indicate that the \textbf{SMSE-OR algorithm} performs well across most datasets, consistently achieving lower average testing prediction error (PE) compared to the other methods, except for the SCRAP DATA dataset, where its performance lags behind.

    \begin{figure}
        \begin{center}
            \includegraphics[width=4.5in]{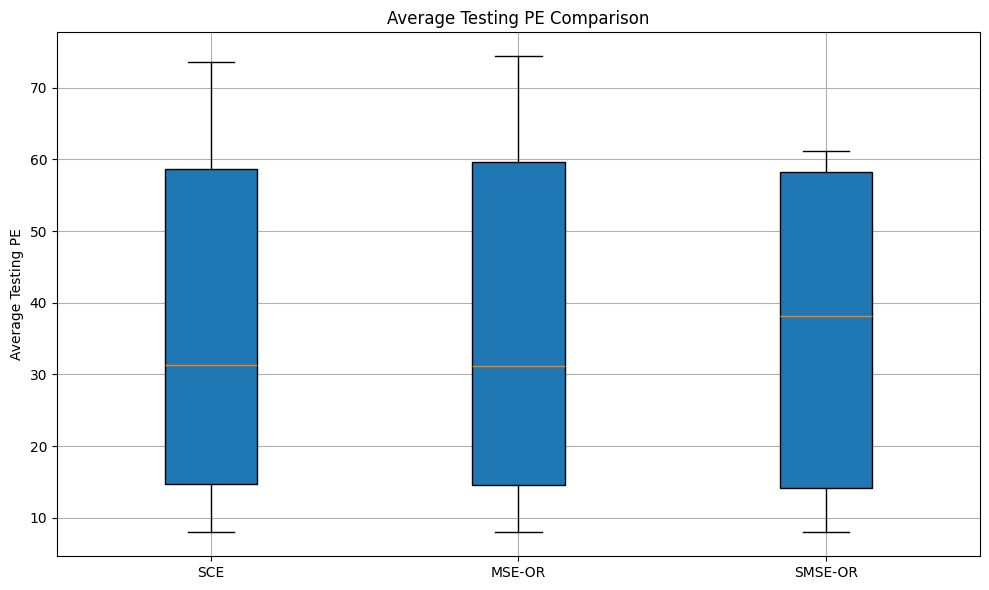}
        \end{center}
        \caption{Average Testing Predictive Error for SCE, MSE-OR, and SMSE-OR Algorithms Across Datasets}
        \label{fig:bp1}
    \end{figure}

    \begin{figure}
        \begin{center}
            \includegraphics[width=4.5in]{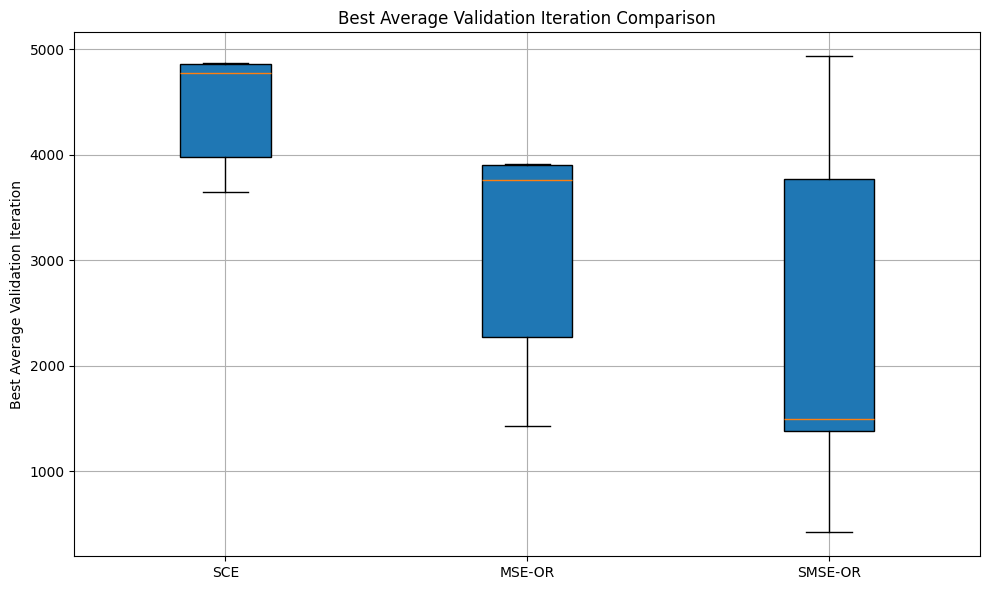}
        \end{center}
        \caption{Best Average Validation Iteration for SCE, MSE-OR, and SMSE-OR Algorithms Across Datasets}
        \label{fig:bp2}
    \end{figure}

Figure \ref{fig:bp1} and \ref{fig:bp2} present a comparative analysis of the performance metrics across three algorithms—SCE, MSE-OR, and SMSE-OR—using two key indicators: Average Testing Predictive Error (PE) and Best Average Validation Iteration. The box plots for Average Testing PE illustrate that the SMSE-OR algorithm consistently achieves the lowest error across all datasets, suggesting a superior capability in generalization compared to the other methods. In contrast, the MSE-OR algorithm exhibits a higher average testing PE, particularly for the CIFAR10 and SVHN datasets, indicating potential challenges in handling more complex data distributions. Additionally, the Best Average Validation Iteration results reveal notable variations in performance across the algorithms. While SCE maintains relatively high validation iterations across most datasets, the MSE-OR algorithm shows a significantly lower average for certain datasets, which may indicate inefficiencies in model training or validation processes. Overall, these findings underscore the effectiveness of the SMSE-OR algorithm in achieving lower predictive errors and suggest areas for improvement in the other algorithms, particularly in complex data environments.

    \begin{figure}
        \begin{center}
            \includegraphics[width=4.5in]{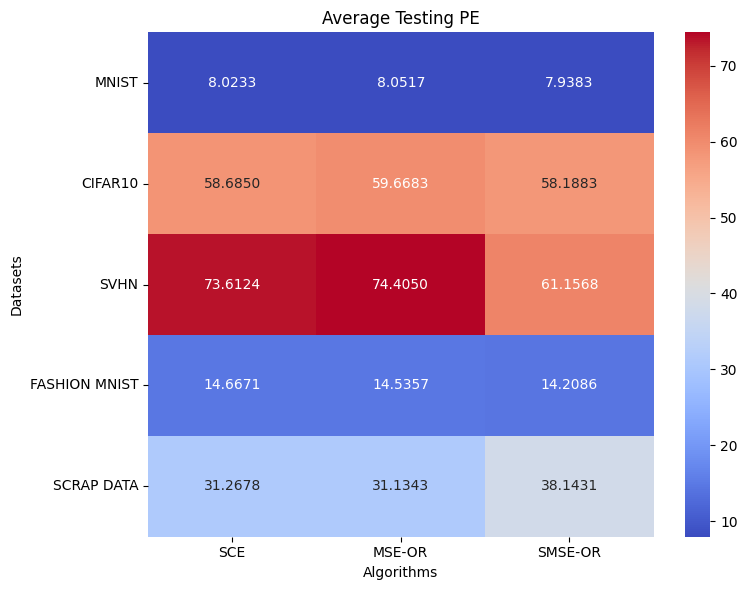}
        \end{center}
        \caption{Average Testing Predictive Error Across Algorithms and Datasets}
        \label{fig:hm1}
    \end{figure}

    \begin{figure}
        \begin{center}
            \includegraphics[width=4.5in]{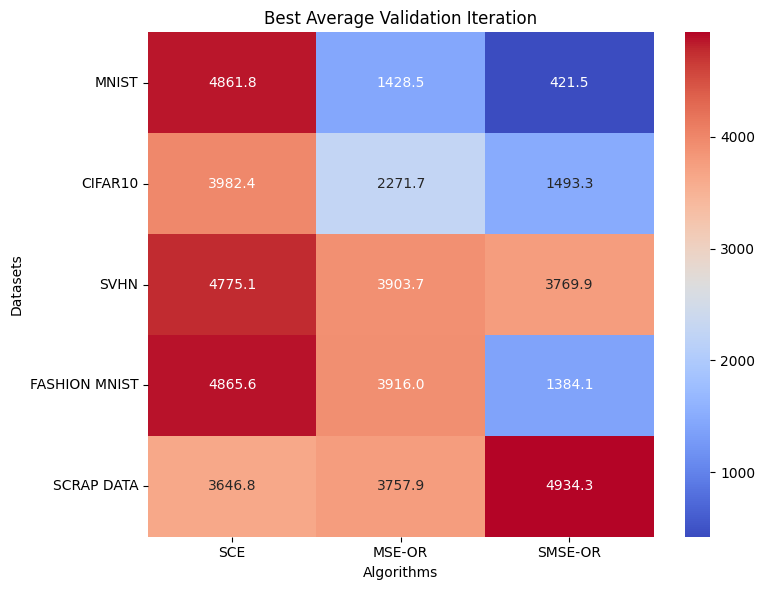}
        \end{center}
        \caption{Heatmap of Best Average Validation Iteration Across Algorithms and Datasets}
        \label{fig:hm2}
    \end{figure}

The heatmap in Figure \ref{fig:bp1} represents the Average Testing Predictive Error (PE) across the three algorithms—SCE, MSE-OR, and SMSE-OR—provides a visual comparison of their performance across various datasets. Each cell in the heatmap is annotated with the corresponding average testing PE value, facilitating an immediate understanding of how each algorithm performs. Notably, the SMSE-OR algorithm consistently demonstrates lower average testing PE values across all datasets, indicating its superior performance in minimizing prediction errors. In contrast, the MSE-OR algorithm exhibits higher testing PE values, particularly with the SVHN dataset, which suggests potential limitations in its capacity to generalize effectively. This visualization succinctly highlights the differences in algorithm effectiveness, guiding future research towards adopting the SMSE-OR approach for better predictive performance. Similarly, the heatmap in Figure \ref{fig:bp2} represents the Best Average Validation Iteration illustrates the number of validation iterations required by each algorithm to achieve optimal performance across the datasets. The annotated values allow for easy comparison of how efficiently each algorithm converges during training. The SCE algorithm generally requires a higher number of iterations to reach its best average validation performance, especially for the MNIST and FASHION MNIST datasets. Conversely, the SMSE-OR algorithm achieves its best average validation iteration with relatively fewer iterations in the majority of datasets, signifying a more efficient training process. These findings suggest that while the SMSE-OR algorithm not only excels in predictive accuracy, it also demonstrates greater efficiency in model training, making it a favorable choice for future implementations in machine learning tasks.

    \section{Discussion}
    
    This study explored the use of MSE combined with a sigmoid activation function for classification tasks. Experiments conducted with a linear classifier highlighted the fundamental interaction between the objective function and activation mechanism, suggesting a unified framework that bridges classification and regression in neural networks.
    
    \begin{enumerate}
        \item Initial results suggest that MSE with sigmoid may exhibit greater robustness to noise or mislabeled data compared to traditional cross-entropy-based methods. This potential advantage calls for further evaluation in real-world scenarios with imperfect datasets.
    
        \item The study reveals distinct differences in the learning dynamics when compared to categorical cross-entropy. Future work should involve a detailed analysis of convergence behavior and generalization, especially in the context of more complex deep learning architectures.
    
        \item The findings emphasize the need for a deeper theoretical understanding of how classification and regression tasks relate within neural networks, particularly in terms of learning trajectories and optimization dynamics.
    
    \end{enumerate}

    \section{Conclusion}
    
    This research presents an innovative approach to neural network design by proposing the use of MSE with sigmoid activation for classification tasks. While traditional models typically use categorical cross-entropy with softmax, our proposed method offers promising alternatives. By providing a framework that unifies classification and regression, this work broadens the theoretical foundations of deep learning. The initial results suggest that MSE with sigmoid may offer increased robustness to noise and data imperfections, making it a potential solution for developing more resilient classifiers in real-world applications with suboptimal data conditions. These findings challenge established norms and prompt a re-evaluation of the theoretical relationship between classification and regression in neural networks, potentially revealing new insights into learning dynamics. This study serves as a catalyst for rethinking conventional neural network paradigms, moving beyond cross-entropy towards more versatile and theoretically informed models. The exploration of MSE with sigmoid activation represents a meaningful step toward deeper understanding and broader applicability of neural network optimization principles.

    \section{Future Work and Open Questions}
    
    This study opens new avenues for further exploration. Evaluating the performance of MSE with sigmoid activation in advanced deep learning architectures (e.g., CNNs, RNNs, Transformers) across different domains is essential to determine its broader applicability. A comprehensive theoretical analysis of convergence and generalization properties for MSE with sigmoid-based classifiers is necessary to better understand their strengths and limitations. Additionally, designing specialized regularization techniques tailored for MSE and sigmoid could further enhance model robustness and generalization. Extensive benchmarking on diverse, real-world datasets will be crucial for a conclusive evaluation of this approach. Investigating the role of MSE with sigmoid in explainable AI may also offer insights into the decision-making processes of neural networks. Several open questions have emerged from this research:
    
    \begin{enumerate}
        \item How does the convergence of MSE with sigmoid-based classifiers compare to that of traditional cross-entropy methods? Does it result in different optima, and what theoretical guarantees can be established?
    
        \item Does MSE with sigmoid foster distinct generalization properties compared to cross-entropy? Are there specific scenarios or dataset characteristics where it excels or shows weaknesses?
    
        \item Can a formal theoretical framework be developed to explain the observed ability of MSE with sigmoid to integrate classification and regression tasks? What are the fundamental mathematical relationships?
    
        \item Are existing optimization algorithms (e.g., Adam, SGD) effective for MSE with sigmoid, or is there a need for customized optimizers to fully exploit this approach?
    
        \item What new regularization methods, designed specifically for MSE with sigmoid, could enhance model robustness and generalization?
    
        \item How does MSE with sigmoid scale to complex architectures like CNNs, RNNs, and Transformers? Are there any architectural adjustments that enhance its effectiveness?
    
        \item How does MSE with sigmoid perform under different types and levels of noise (e.g., label noise, feature noise)? Are there measurable advantages over cross-entropy-based methods?
    
        \item How does MSE with sigmoid compare to state-of-the-art classification methods on a wide range of real-world datasets across different domains (e.g., vision, NLP)?
    
        \item Does the use of MSE with sigmoid affect a model's vulnerability to adversarial attacks?
    
        \item How do decision boundaries formed by MSE with sigmoid differ from those of cross-entropy models? Do these differences impact interpretability?
    
        \item Can established feature attribution methods (e.g., saliency maps) be effectively applied to MSE with sigmoid-based models, or do they require adaptation?
    \end{enumerate}

\bibliographystyle{unsrt}
\bibliography{reference}

\appendix

\end{document}